\documentclass[10pt,twocolumn,letterpaper]{article}

\usepackage{wacv}
\usepackage{times}
\usepackage{epsfig}
\usepackage{graphicx}
\usepackage{amsmath}
\usepackage{amssymb}

\usepackage{breqn} 
\usepackage{verbatim}
\graphicspath{{images/}}
\usepackage{subcaption} 

\wacvfinalcopy 

\def\wacvPaperID{403} 

\ifwacvfinal\fi
\begin{document}

\title{Neural Signatures for Licence Plate Re-identification}

\author{Abhinav Kumar\thanks{These two authors contributed equally}\\
Conduent Labs\\
{\tt\small abhinav.kumar@conduent.com}
\and
Shantanu Gupta\footnotemark[1] \footnotemark[2]\\
University of Wisconsin-Madison\\
{\tt\small sgupta226@wisc.edu}
\and
Vladimir Kozitsky\thanks{The work was carried out when they were at Conduent Labs}\\
Palo Alto Research Centre \\
{\tt\small vladimir.kozitsky@parc.com}
\and
Sriganesh Madhvanath \\
Conduent Labs\\
{\tt\small srig@acm.org}
}
\maketitle

\ifwacvfinal\thispagestyle{empty}\fi

\begin{abstract}
    The problem of vehicle licence plate re-identification is generally considered as a one-shot image retrieval problem. The objective of this task is to learn a feature representation (called a \emph{``signature''}) for licence plates. Incoming licence plate images are converted to signatures and matched to a previously collected template database through a distance measure. Then, the input image is recognized as the template whose signature is ``nearest'' to the input signature. The template database is restricted to contain only a single signature per unique licence plate for our problem.

    We measure the performance of deep convolutional net-based features adapted from face recognition on this task. In addition, we also test a hybrid approach combining the Fisher vector with a neural network-based embedding called ``f2nn'' trained with the Triplet loss function. We find that the hybrid approach performs comparably while providing computational benefits. The signature generated by the hybrid approach also shows higher generalizability to datasets more dissimilar to the training corpus.\\

    \normalfont{\textbf{Keywords:} Signature Matching, Optical Character Recognition, Fisher Vectors, Neural Networks, Triplet Loss, Transfer Learning, Image Retrieval, Recommendation.}
\end{abstract}

\section{Introduction \label{sec:Introduction}}
    \emph{Automatic vehicle identification} is a common problem in designing intelligent transportation systems, such as automatic tolling systems and automatic parking ticket administering systems. A desirable modality to use in such problems is CCTV-grade camera footage, as dedicated machinery such as transponders and receivers often proves quite costly. Using cameras also opens up opportunities for applications using phone cameras, e.g.  to aid law enforcement officers with access to vehicle registration information.
    
    The canonical vehicle identification problem in the context of images is \emph{licence plate recognition} (LPR), is thought of as two problems in sequence: localizing a licence plate in a full-field image (in general) and then recognizing the plate from the extracted region of interest (ROI) in the image. This work focuses on the second problem; for plate localization the reader is directed to \cite{redmon2016you}.
    
    \emph{Optical character recognition} (OCR) is the \emph{de facto} approach to licence plate recognition. Its results are fairly good in practice but it occasionally fails to read the letters with a high enough confidence (due to difficulties in separating text cleanly from patterned backgrounds, symbol variations across images, and occasional similarities in the shapes of different symbols in some fonts). However, for commercially viable LPR systems, it is prudent to explore alternative approaches which improve overall recognition performance when used in tandem with OCR, as seen in many other classification problems when multiple models are aggregated. One such approach is vehicle re-identification, which is restricted in scope compared to general licence plate recognition but provides a solid alternative in scenarios where individual vehicles are observed repeatedly.
    

    \subsection{Problem Overview \label{sec:Introduction:overview}}
        The problem we attempt to address in this work is \emph{one-shot licence plate re-identification}, that is, we need to learn a feature representation (aka signature) for licence plates which is general enough that once we see a single labelled image of a licence plate, we recognize another image of that licence plate reliably -- an ideal system would provide an interpretable confidence measure as well. Different captures of the same license plate display variations such as differing camera angles, lighting variations, occlusion, noise such as shadow, and different ROI crops: we would like our feature representations to be invariant to such variations. Thus, the goal is to \textit{learn a signature generation model} that generates signatures that are used for re-identifying new license plate images with high accuracy. The size of the signature is also crucial since the matching time for a new license image in the database of images is a function of signature dimension in general, and specifically, $O(D)$ for $D-$dimensional cosine similarity-based matching. 
        
        The rest of this paper is organized as follows: section \ref{sec:related_work} surveys various signature generation methods and the loss functions used in training them. Section \ref{sec:f2nn} talks about the proposed approach \emph{f2nn}. 
        The data used for training, validation and benchmarking is described in Section \ref{sec:data_splits}. Section \ref{sec:expt} discusses the experimental setup while Section \ref{sec:results} focuses on results and the discussions that follow. 
        It also suggests possible directions this work can lead towards.

\section{Related Work \label{sec:related_work}}
    \cite{rodriguez2012data},\cite{serrano2013methods} demonstrate the viability of using a feature extraction pipeline as a \emph{signature} extractor, from which a licence plate's feature representation can be retrieved in cases when a particular plate has been seen before.

    As previously mentioned, the task is to learn a signature generation model. There are broadly two classes of approaches to learning signature generation models: unsupervised and supervised. Unsupervised approaches do not use the labeled data while supervised approaches use labels of the data. Some hybrid methods also exist, which combine these approaches. Loss functions play a crucial role in the learning process.

    \subsection{Signature Generation \label{sec:related_work:sig_gen}}
        The earliest approaches to generate image signatures for classification was using bag-of-visual-words (BOV) histograms \cite{csurka2004visual}. However, the signature generation is a lossy process \cite{boiman2008defense} and is not scalable to thousands of images. The most popular unsupervised signature model in the literature is the Fisher Vector and its variants \cite{perronnin2007fisher}, \cite{perronnin2010improving}. The Fisher Vector, introduced by \cite{perronnin2007fisher}, is a vector that expresses a particular data point in terms of its relation to a statistical model (concretely, to the derivatives with respect to the model parameters, computed at the data point). The statistical model is fit to an unlabelled training corpus of similar data. In our case, the model we use is a mixture of Gaussian probability distributions, fit to (roughly) a post-processed dense SIFT descriptor \cite{lowe2004distinctive} extracted from the licence plate ROIs of a corpus of licence plate images collected from on-road operations. Notably, the model is trained at different spatial scales and aggregated over different image regions via Spatial Pyramid Pooling. \cite{perronnin2010improving}, \cite{lazebnik2006spatialpyr}.
        
        Convolutional Neural Networks (CNNs) are the most popular supervised models in computer vision in recent years. CNNs have overtaken Fisher Vectors in many computer vision tasks such as image recognition \cite{krizhevsky2012imagenet}, \cite{szegedy2015going}, \cite{He2015} and face recognition \cite{SchroffEtAl15}, \cite{ParkhiEtAl15}.

        Despite their success in several computer vision tasks, CNNs need a high forward pass time which may not be very suitable for many business requirements. Also, CNNs lack geometric invariance \cite{gong2014multi}. It is possible to combine multiple approaches to creating hybrid signature models \cite{simonyan2013deep}, \cite{sydorov2014deep} \cite{PerronninLarlusCVPR15}. \cite{simonyan2013deep} carried out dimensionality reduction of Fisher Vectors. \cite{sydorov2014deep} jointly learned the SVM classifier and the GMM visual vocabulary. Perronnin et al \cite{PerronninLarlusCVPR15} trained a shallow-net (a single- or a 2-layer network) with Fisher Vectors as inputs and softmax loss to obtain a lower dimensional embedding from the original Fisher vectors. Our approach is based on this work but employs loss functions more attuned for one shot learning. 

            \begin{figure*}[!t]
                \begin{center} \includegraphics[width=0.8\linewidth,keepaspectratio]{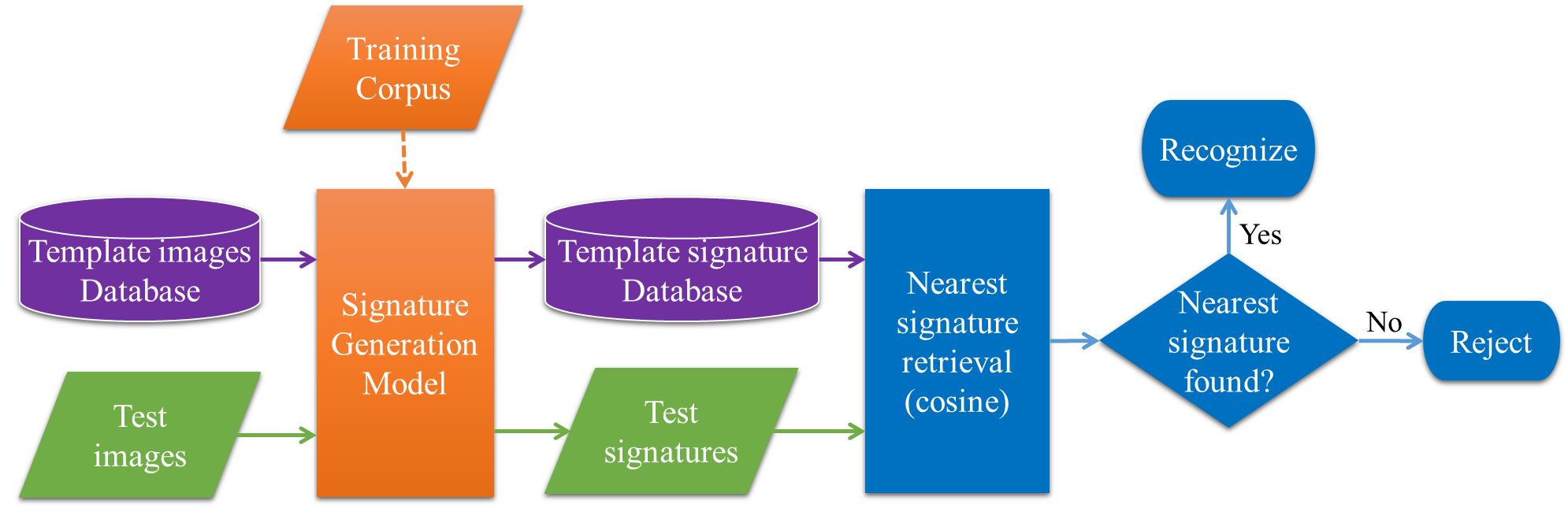}
                    \caption{Flow chart of the benchmarking procedure.}
                    \label{fig:flow}
                \end{center}
            \end{figure*}
    \subsection{Loss functions \label{sec:related_work:loss_fns}}
        Multiple loss functions apart from the softmax function have been proposed in the literature for the purposes of training discriminative embeddings, such as the center loss \cite{Wen2016}, the Siamese loss function \cite{chopra2005learning}, the contrastive loss \cite{hadsell2006dimensionality}, the triplet loss (\cite{SchroffEtAl15}) and the quadruplet loss (\cite{chen2017beyond}) among others.

        \subsubsection{Center loss \label{sec:related_work:loss_fns:center_loss}}
            The center loss \cite{Wen2016}, in practice, is a regularized softmax loss, which explicitly penalizes distance from the centroid of the feature representations of the data points of a given class, or equivalently, the intra-class variance.

        \subsubsection{Triplet loss \label{sec:related_work:loss_fns:triplet_loss}}
            The triplet loss \cite{SchroffEtAl15} is seen as a generalization of the Siamese network loss\cite{chopra2005learning}. It applies on a triplet $(x_a, x_p, x_n)$ where $x = x(i)$ for $i \in \{a, p, n\}$ refers to the CNN embedding given images $a$, $p$, and $n$ such that $a$ and $p$ belong to the same class and $n$ comes from a different class. Moreover, the embeddings are expected to be normalized -- we can use either $L_2$-normalization or batch-normalization \cite{ioffe2015batch} for this purpose. As both techniques performed similarly in our tests, we only consider results from batch-normalized embeddings for the rest of the paper, as that is the more common normalization used in practice. As an aside, we note that, in conformance with \cite{DBLP:journals/corr/WangXCY17}, we find that scaling the normalization layer output to adjust the norm of the $L_2$-normalized embedding is crucial in getting the softmax loss to reduce at all -- we use a norm of $\sqrt{d}$ for a $d$-dimensional embedding for our experiments.
            
            The triplet loss for one triplet $(x_a, x_p, x_n)$ is computed as
            \begin{equation}\label{eq:Trip}
            \begin{split}
            L(x_a, x_p, x_n) & = \max(||x_a - x_p||_2^2 - ||x_a - x_n||_2^2 + \alpha, 0) \\
                             & = \left[||x_a - x_p||_2^2 - ||x_a - x_n||_2^2 + \alpha\right]_+
            \end{split}
            \end{equation}
            
            The loss is added up when there are multiple images in a minibatch, and many combinations of data points are taken to form triplets. If the loss and the number of active triplets (which have a non-zero loss value) reduces to zero during training, that means that the training data has been separated in the embedding space with a margin of $\alpha$.
        
        \subsubsection{Quadruplet loss \label{sec:related_work:loss_fns:quadruplet_loss}}
            Proposed in \cite{chen2017beyond}, this loss further generalizes the triplet loss. 
            It applies on a quadruplet $(x_a, x_p, x_n,  x_{n_2})$ where $x = x(i)$ for $i \in \{a, p, n, n_2\}$ refers to the embedding computed by the CNN model when given images $a$, $p$, $n$ and $n_2$ such that $a$ and $p$ belong to the same class, $n$ comes from a different class and $n_2$ comes from a class other than $a$ and $n$.

            \begin{dmath}\label{eq:quad}
            L(x_a, x_p, x_n, x_{n_2}) =
                \left[||x_a - x_p||_2^2 - ||x_a - x_n||_2^2 + \alpha_1\right]_+ \\
                + \left[||x_a - x_p||_2^2 - ||x_n - x_{n_2}||_2^2 + \alpha_2\right]_+
            \end{dmath}

            The metric could itself be learned using the training set but we stick to the $L_2$ norm since that would not degrade the performance on the transfer set and also simplify the training process.

\section{Proposed Solution: \emph{f2nn} \label{sec:f2nn}}
    \subsection{Procedure}\label{sec:f2nn:procedure}
    The images are passed through the signature matching module which finds the nearest template signature to the test image's signature (using the cosine distance) and assigns the test plate to the corresponding vehicle. The entire flow is shown in fig. \ref{fig:flow}. The requirement of the system is robustness to both type 1 and type 2 errors : it should reject plates which are not present in the system, and should not reject plates which are already in the system. Depending on the application, we assign higher priority to minimizing either type 1 or type 2 errors -- in this work they are treated equally.

\begin{figure}[!t]
    \begin{center}
        \includegraphics[width=0.48\linewidth]{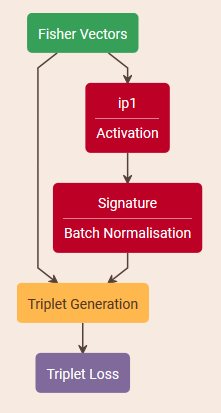}
        \caption{\emph{f2nn} Architecture}
        \label{fig:f2nn_arch}
    \end{center}
    \end{figure}

    Building on previous work in \cite{PerronninLarlusCVPR15}, we use a shallow (2-layer network) with the Fisher vector as input to obtain a much lower dimensional embedding from the original 8192-dimensional Fisher vector. This architecture, henceforth called \emph{f2nn}, is shown in fig \ref{fig:f2nn_arch}.

\section{Data Splits \label{sec:data_splits}}

    \subsection{Training corpus \label{sec:data_splits:training_corpus}}
        We use a set of 120,252 images, from 49,872 unique vehicles in the US, as a corpus for training the models described here. 
    
    \subsection{Validation Data \label{sec:data_splits:validation_data}}
        In addition, we keep aside 24,191 images from 10,000 other unique American vehicles to use for validation and model hyperparameter tuning. 

    \subsection{Benchmarking Data \label{sec:data_splits:benchmark_data}}
\begin{figure}[!t]
        \centering
            \begin{subfigure}[b]{0.48\linewidth}
                \centering
                \includegraphics[width=0.45\linewidth]{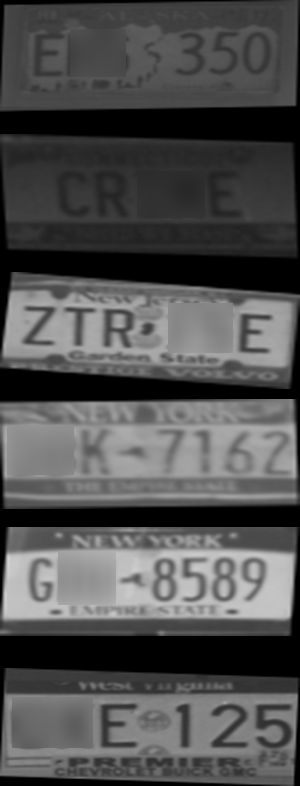}
                \caption{US license plates}
                \label{fig:mta}
            \end{subfigure}
            \begin{subfigure}[b]{0.48\linewidth}
                \centering
                \includegraphics[width=0.45\linewidth]{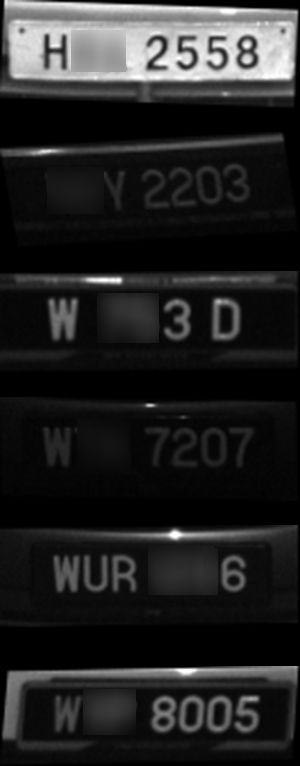}
                \caption{Malaysian license plates}
                \label{fig:mhi}
            \end{subfigure}
            \caption{Sample license plates from the two datasets. Some letters of license plates have been redacted to preserve privacy.}
            \label{fig:mta_mhi}
        \end{figure}

        To benchmark our models, we use two datasets of cropped licence plate image ROIs:
        \begin{itemize}
        	\item 25,934 Malaysian licence plate ROIs from 11,200 unique vehicles
        	\item 24,249 American licence plate ROIs from 10,000 unique vehicles
        \end{itemize}
        The two datasets (training and validation), and the benchmark set from the US described previously are part of the same overall US dataset, and hence contain fairly similar images; however, the sets of licence plates taken in the three sets are completely disjoint, so a model trained to recognize just some particular licence plates cannot be used. The Malaysian benchmark dataset has images which look somewhat different in appearance, and can therefore test the generalization capabilities of the models we train to a little larger extent. A few sample images from the two datasets have been shown in fig. \ref{fig:mta_mhi}.

\section{Experimentation \label{sec:expt}}

    \subsection{Models \label{sec:expt:models}}
        
        \subsubsection{Fisher vectors \label{sec:expt:models:fisher}}
            The Fisher vectors we use are derived from a closed implementation which we could not access at the time of writing, and are therefore treated as a black-box feature representation of the licence plate images. The FV model was trained on an offline corpus of US licence plate ROIs, which was also inaccessible to us at the time of writing. Once the Fisher vectors are extracted from the training images, all further finetuning and benchmarking is performed on the common dataset described in section \ref{sec:data_splits}.
            
            \begin{figure}
                \begin{center}
                    \includegraphics[width=0.6\linewidth]{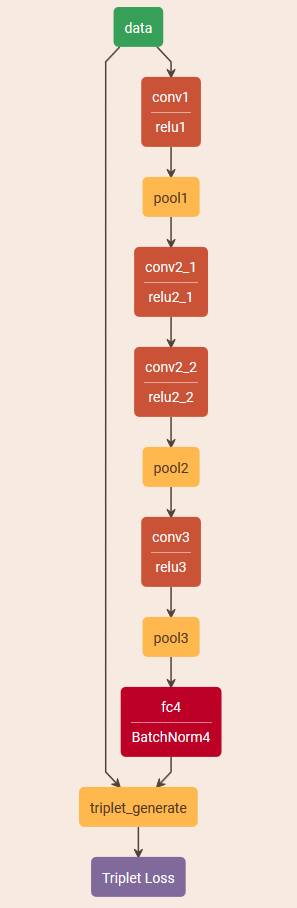}
                    \caption{TFS Architecture}
                    \label{fig:tfs}
                \end{center}
                \end{figure}
            
        \subsubsection{CNNs \label{sec:expt:models:cnn}}
            We use some commonly known CNN architectures to compare the proposed embedding with:
            \begin{itemize}
            	\item The VGG-Face model described in \cite{ParkhiEtAl15} (pre-trained weights available online).
            	
            	
            	\item The ResNet-50 architecture from \cite{He2015}.
            \end{itemize}
            
            In addition, we also use a smaller VGG-like architecture called ``TFS'' which we train from scratch on the training corpus. Its architecture is shown in fig \ref{fig:tfs}. This network is quite fast to train, and the motivation for using a small architecture is that a licence plate dataset has much less native variation and input resolution than ImageNet, so a smaller model might suffice. 

        \subsubsection{Loss functions \label{sec:expt:models:loss}}
            In addition to the loss functions described in section \ref{sec:related_work:loss_fns}, we also use a squared-error (Euclidean) loss function in an autoencoder architecture for the Fisher vector embedding. The encoding part of the autoencoder network is kept the same as that for the models used with the other loss functions (i.e., the triplet loss, among others). For the center loss, we use our own implementation using the details given in \cite{Wen2016}.

    \subsection{Preprocessing \label{sec:expt:preprocessing}}
        The Fisher vector pipeline is agnostic to the exact image size and is extracted from multiple scales of input image, and therefore we use the original images to derive Fisher vectors. No preprocessing is applied.
        
        For training the CNNs, we resize the images to 224-by-224 using bicubic interpolation, taking care to preserve the image aspect ratio. Non-square images are resized such that the larger dimension becomes 224; this ensures that no image content is lost in the preprocessing stage. The remaining space is padded with zeros.
        
        We also attempted some image normalization using histogram equalization, but as results were inconclusive either way, we elect to remove it from the results presented here.
        
    \subsection{Training \label{sec:expt:training}}
        For all experiments we use a couple of NVIDIA Tesla K80 GPUs, with around 12 GB of VRAM. We used Caffe \cite{jia2014caffe} and Lua Torch \cite{torch} for all experiments presented here.
        
        Model parameters are seen to take up to 1 GB of VRAM, with the rest of the VRAM filled by layer activations. Typical minibatch sizes for CNN training are 32 to 64 images. As the Fisher vector embedding network is quite shallow and layer activations have fairly small size, we could in turn use much larger batch sizes for training that model -- the typical batch size for \emph{f2nn} is of the order of 1,000.
        
        Taking combinations of data points from a batch to form triplets can be done in multiple ways. We performed this triplet mining procedure in two ways:
            \begin{itemize}
            	\item \textbf{Online triplet mining}: To choose which triplets we form from all the possible ones, we use the \emph{semi-hardest triplet selection} rule, by which we take the triplets which give the highest value for the loss function, which effectively means that we take triplets where the negative examples are the closest possible to the positive example, while still not being closer to the anchor than the positive itself. Equivalently, the selected triplets are the ones which are the least separated. The triplets selected are only called semi-hard because we ignore triplets where the negative example is closer to the anchor than the positive example, as they are said (\cite{SchroffEtAl15}, \cite{hermans2017defense}) to cause ``bad local minima early on in training, specifically it results in a collapsed model (\emph{i.e., $f(x) = 0$})''.
            	
            	\item \textbf{Offline triplet mining}: This proceeds similarly to online triplet mining, except that we initially work using an external feature representation (in our case, the Fisher vector) to generate triplets (with similar triplet numbers -- 1 to 3 positive examples and around 5 negative examples per anchor point) and use the entire dataset to find triplets instead of minibatches (subject to practical constraints -- to reduce computational load we still split the data into around 10 chunks to find triplets). This procedure gives a larger variety in triplets, but requires somewhat more manual effort to perform, without yielding any significant improvement in the results. Therefore we restrict the results presented in this paper to those obtained through online triplet mining.
            \end{itemize}
        
        For the triplet loss, we create the minibatches such that there are around 2 examples of each class in a minibatch. We then select around 5 negative examples per anchor point; yielding, for example, a batch of 5000 triplets, if we take 1 positive example and 5 negative examples per anchor point using a batch size of 1000. Quadruplet loss training proceeds similarly, but with one more dimension selected along.
    
    \subsection{Metrics \label{sec:expt:metrics}}
        We evaluate the signatures based on the following measures, for which we use the notation in table \ref{tab:metrics}.
        
        \begin{table}[!h]
        \caption{Categories of labels assigned to images of the two transfer set.}\label{tab:metrics}
        \begin{center}
        \begin{tabular}{|c|c|c|c|}
        \hline
        ~ & Correct & Wrong & Rejected \\
        ~ & Label & Label & ~ \\
        \hline
        Present in Template & $n_1$ & $n_2$ & $n_3$\\
        Absent in Template  & $n_4=0$ & $n_5$ & $n_6$\\
        \hline
        \end{tabular}
        \end{center}
        \end{table}
 It is clear that $n_4$ has to be $0$, as a plate absent in the template set cannot be recognized correctly.
        
        \begin{itemize}
            \item \textbf{Yield (or recall):} This is computed as:
                \begin{equation}
                    \text{Yield} = \frac{n_1 + n_2}{n_1 + n_2 + n_3}
                \end{equation}
            \item \textbf{False positive rate:} which is defined as
                \begin{equation}
                    \text{FPR} = \frac{n_5}{n_5 + n_6}
                \end{equation}
            \item \textbf{Accuracy (or precision):} In this work, we define accuracy as follows:
                \begin{equation}
                    \text{Accuracy} = \frac{n_1}{n_1 + n_2}
                \end{equation}
                where we do not take into account false matches, as they are handled separately with the false matching rate. The false matching rate measures type 1 errors, while the yield measures type 2 errors, with the accuracy metric checking whether the plates recognized by the system are actually correct.
        \end{itemize}    
    \subsection{Hyperparameter Tuning for \emph{f2nn} \label{sec:expt:hyperparam}}
    We use the validation data from the American license plate dataset for hyperparameter tuning. The Malaysian dataset tests the generalisability of the signature generation and was not used at all.
    
\begin{figure}[!ht]
        \begin{center}
            \includegraphics[width=1\linewidth]{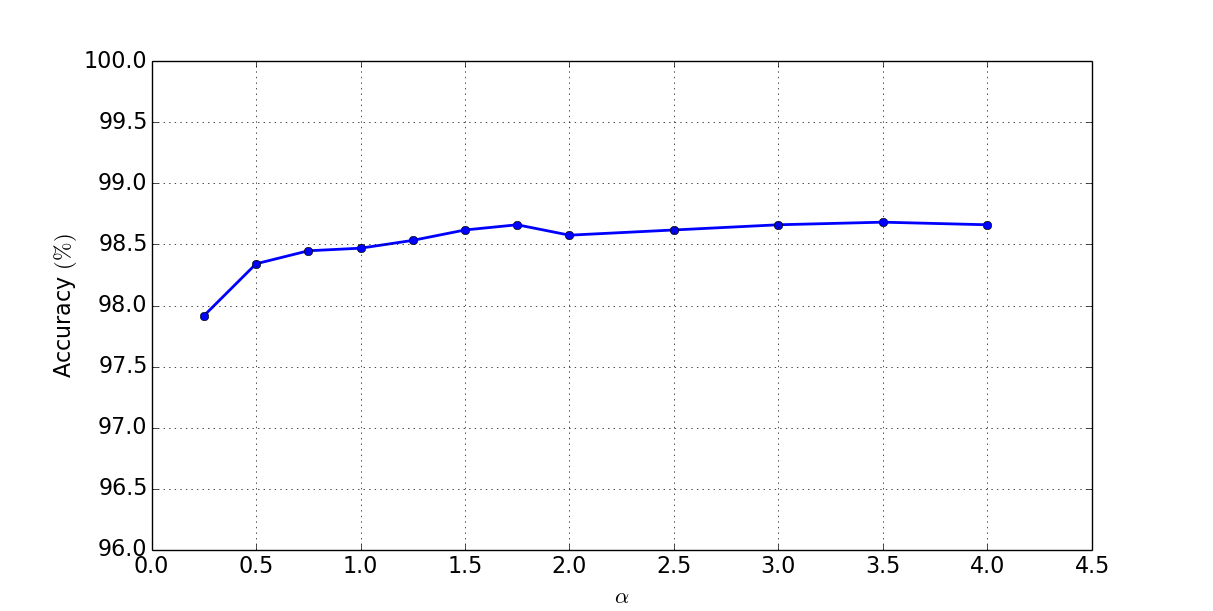}
            \caption{Plot of accuracy vs triplet loss margin $\alpha$ for the American validation dataset}
            \label{fig:acc_alpha}
        \end{center}
        \end{figure}

\begin{figure}[!ht]
        \begin{center}
            \includegraphics[width=1\linewidth]{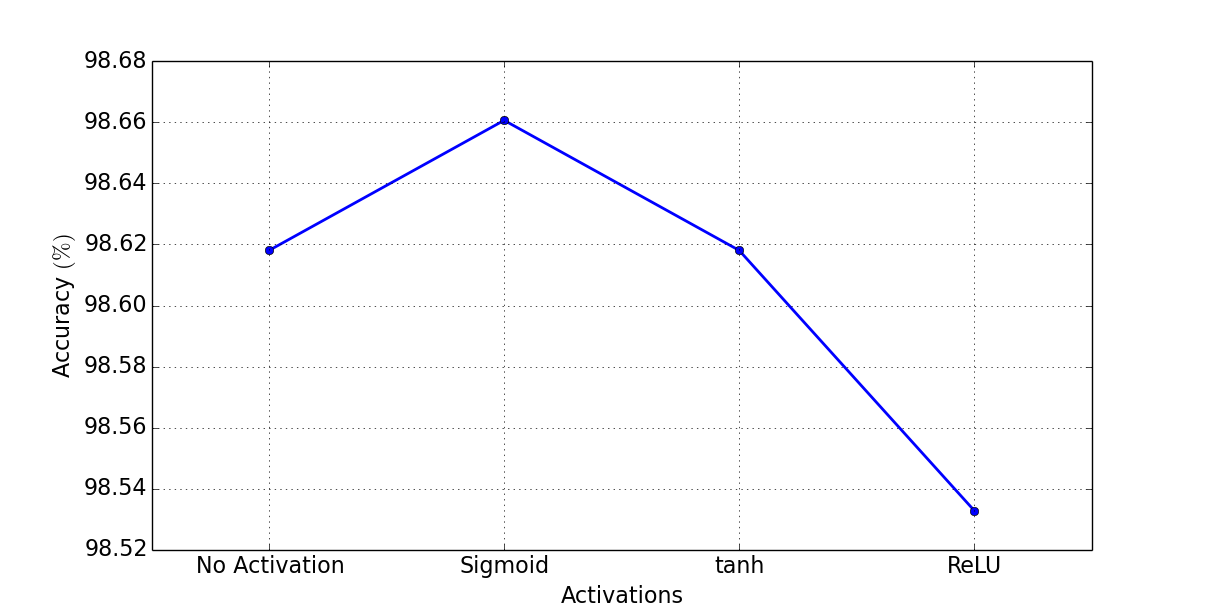}
            \caption{Plot of accuracy vs different activations for the American validation dataset}
            \label{fig:acc_act}
        \end{center}
        \end{figure}

    The validation curves have been shown in figure (\ref{fig:acc_alpha}) to figure (\ref{fig:acc_act}).
        \begin{figure}[!ht]
        \begin{center}
            \includegraphics[width=1\linewidth]{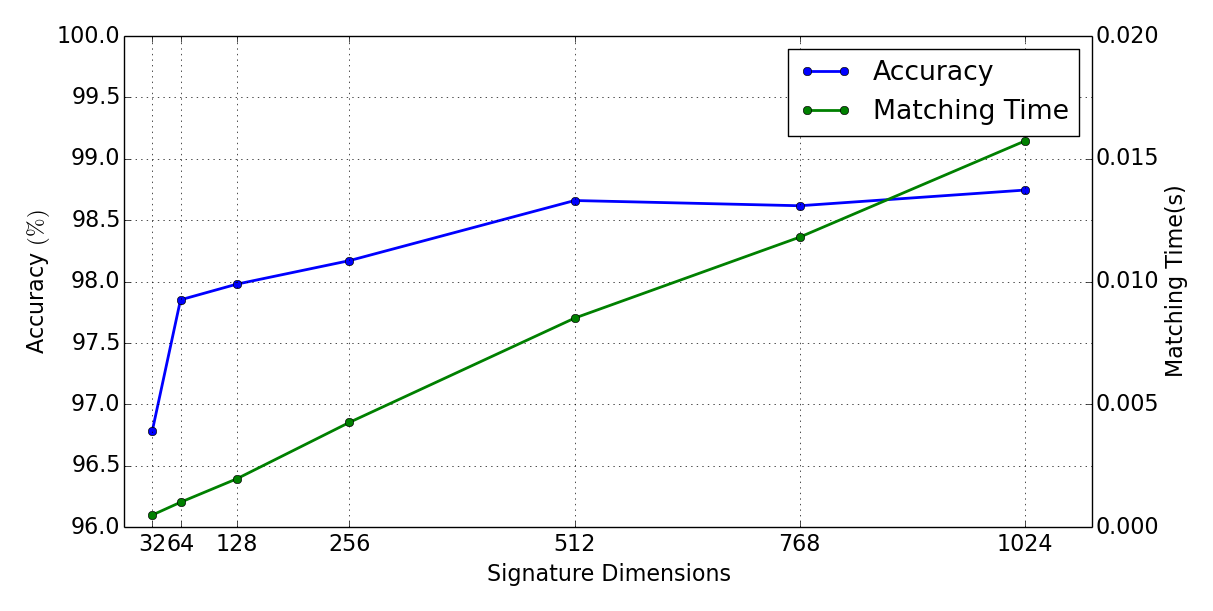}
            \caption{Plot of accuracy and matching time vs Signature dimensions for the American validation dataset}
            \label{fig:acc_sig}
        \end{center}
        \end{figure}
        \begin{figure}[!ht]
        \begin{center}
            \includegraphics[width=1\linewidth]{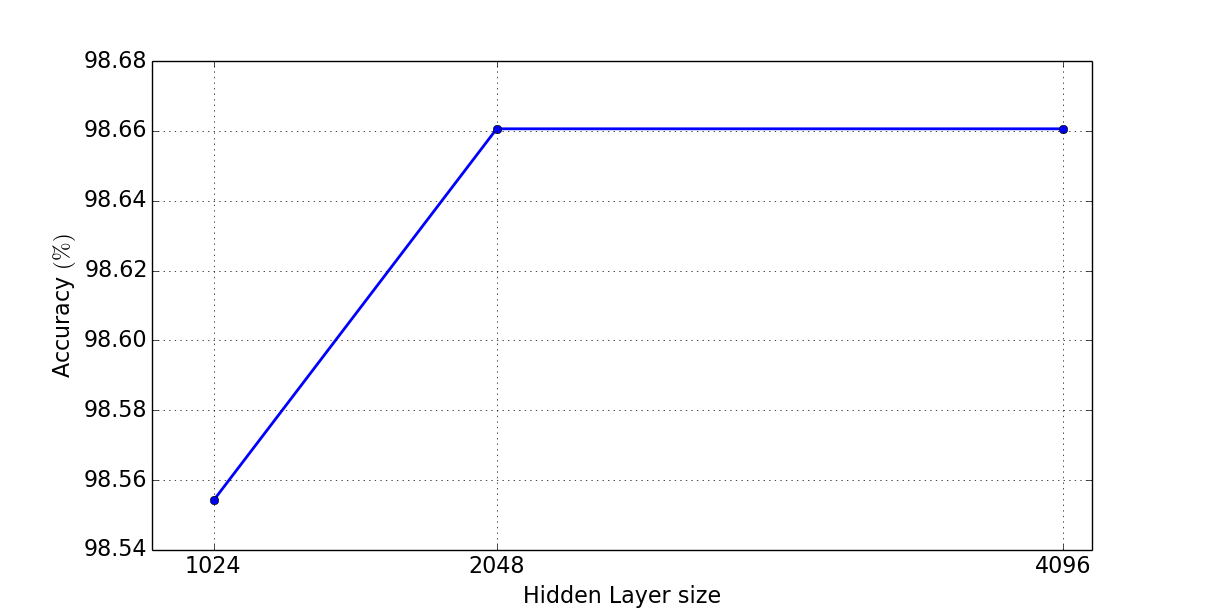}
            \caption{Plot of accuracy vs hidden layer dimensions for the American validation dataset}
            \label{fig:acc_hid}
        \end{center}
        \end{figure}
        
    We chose accuracy as the primary criteria for deciding the hyperparameters. The signature dimensions was chosen to be 512 although the accuracy is less since increasing the dimensions beyond 512 improves the accuracy slightly. We also chose value of $\alpha=1.75$ among other values since smaller values of $\alpha$ ensures better generalizabilty across datasets. The hidden layer was chosen to be of 2048 dimensions and the activation used was sigmoid. Kaiming initialisation
    \cite{he2015delving} was used for initialising the layer weights of the \emph{f2nn} architecture while layer biases were initialised to zero.
    
    We also tried using dropouts \cite{srivastava2014dropout} for the f2nn architecture in fig \ref{fig:f2nn_arch} but that caused the accuracies to drop. So, dropout was not used in the f2nn architecture.
    \subsection{Benchmarking \label{sec:expt:benchmark}}
        We select roughly 60\% of the vehicles (6,693 from Malaysia and 6,006 from the US) to have the signatures from one labelled example each stored in a template array. The 60\% figure matches typical metropolitan daily commuter ratios and is taken to represent the fact that some vehicles will be new for the LPR system, and the signature matching module should recognize that and avoid classifying such images at all.
        
        The feature vectors obtained from both the CNNs and the Fisher vectors (raw and fine-tuned) are tested as described in section \ref{sec:f2nn:procedure}, with the test licence plates held out such that they have never been seen during any phase of training by any of the models we benchmark. To test the generalizability of the system beyond the domain we train it with, we also use the second test set containing images of licence plates from Malaysia, which appear quite different visually compared to American licence plates.
        
        An important point to note is that we perform benchmarking only with one dataset at a time, that is, we do not store the Malaysian images in the template database when testing with the plates from the US, and \emph{vice versa}.
        
        \begin{figure}[!t]
        \begin{center}
            \includegraphics[width=1.1\linewidth]{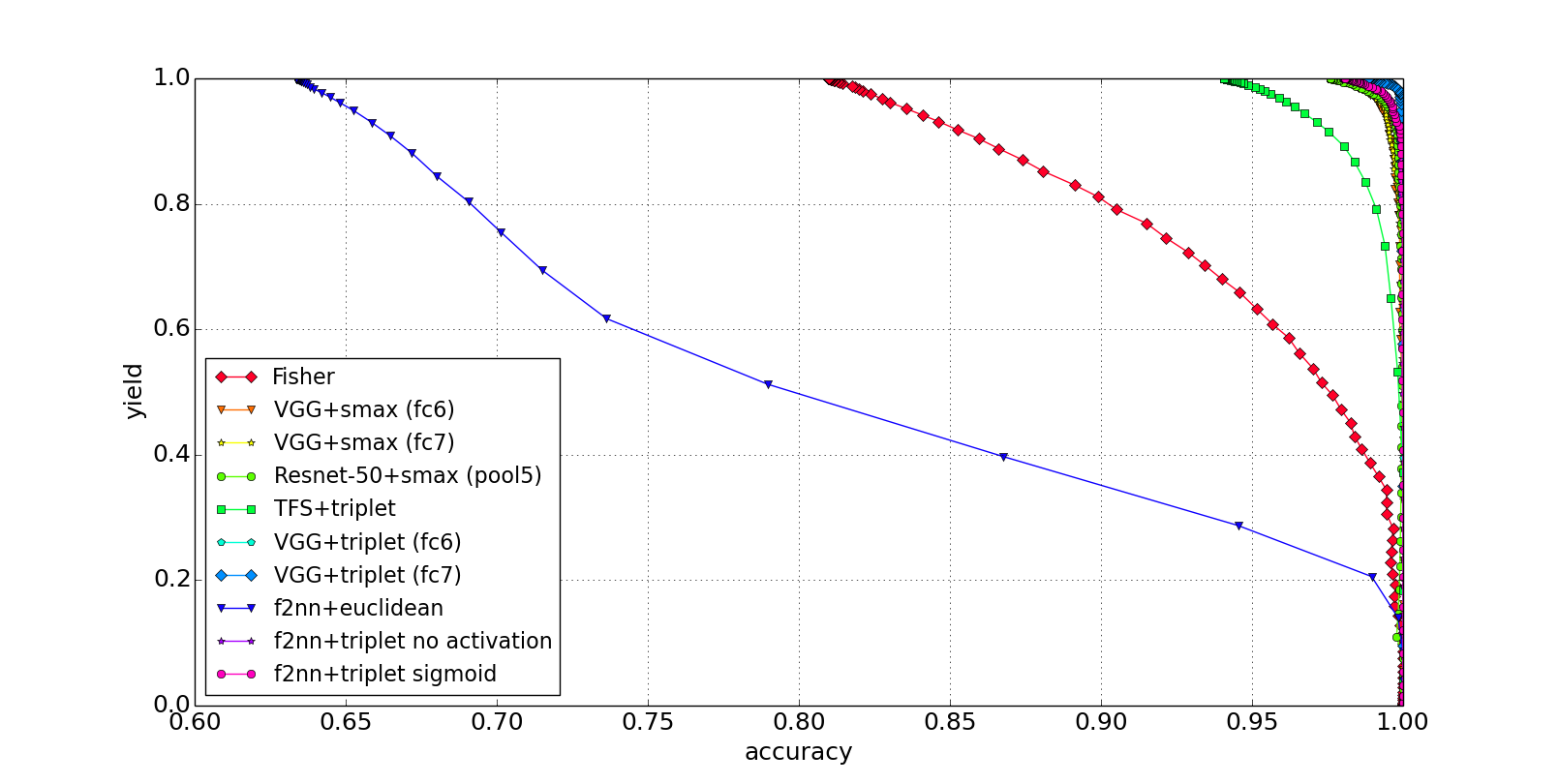}
            \caption{Yield vs Accuracy plot for the American Dataset}
            \label{fig:mta2_acc}
        \end{center}
        \end{figure}
        
        \begin{figure}[!t]
        \begin{center}
            \includegraphics[width=1.1\linewidth]{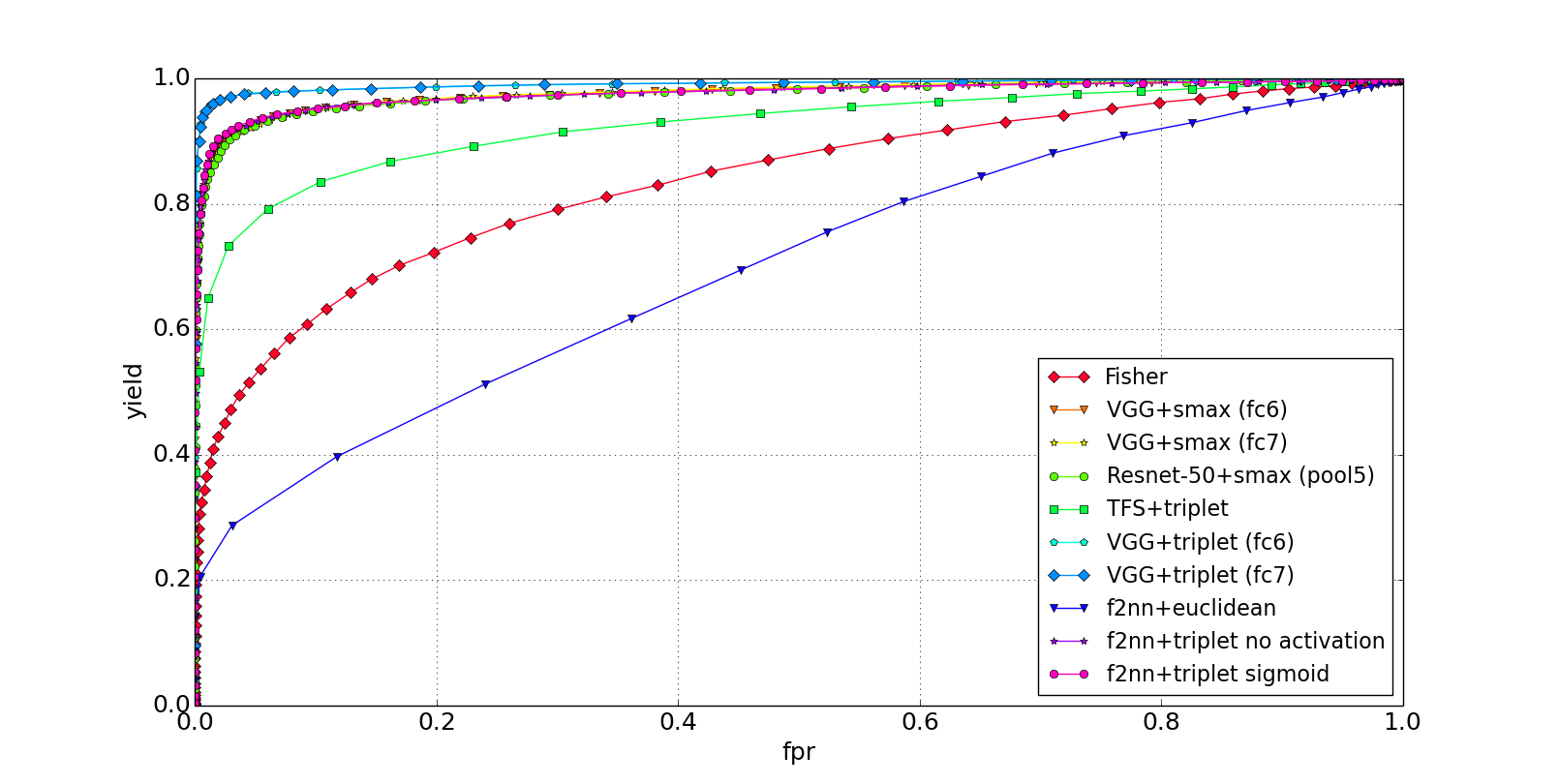}
            \caption{Yield vs FPR plot for the American Dataset}
            \label{fig:mta2_fpr}
        \end{center}
        \end{figure}
    
        \begin{figure}[!t]
        \begin{center}
            \includegraphics[width=1.1\linewidth]{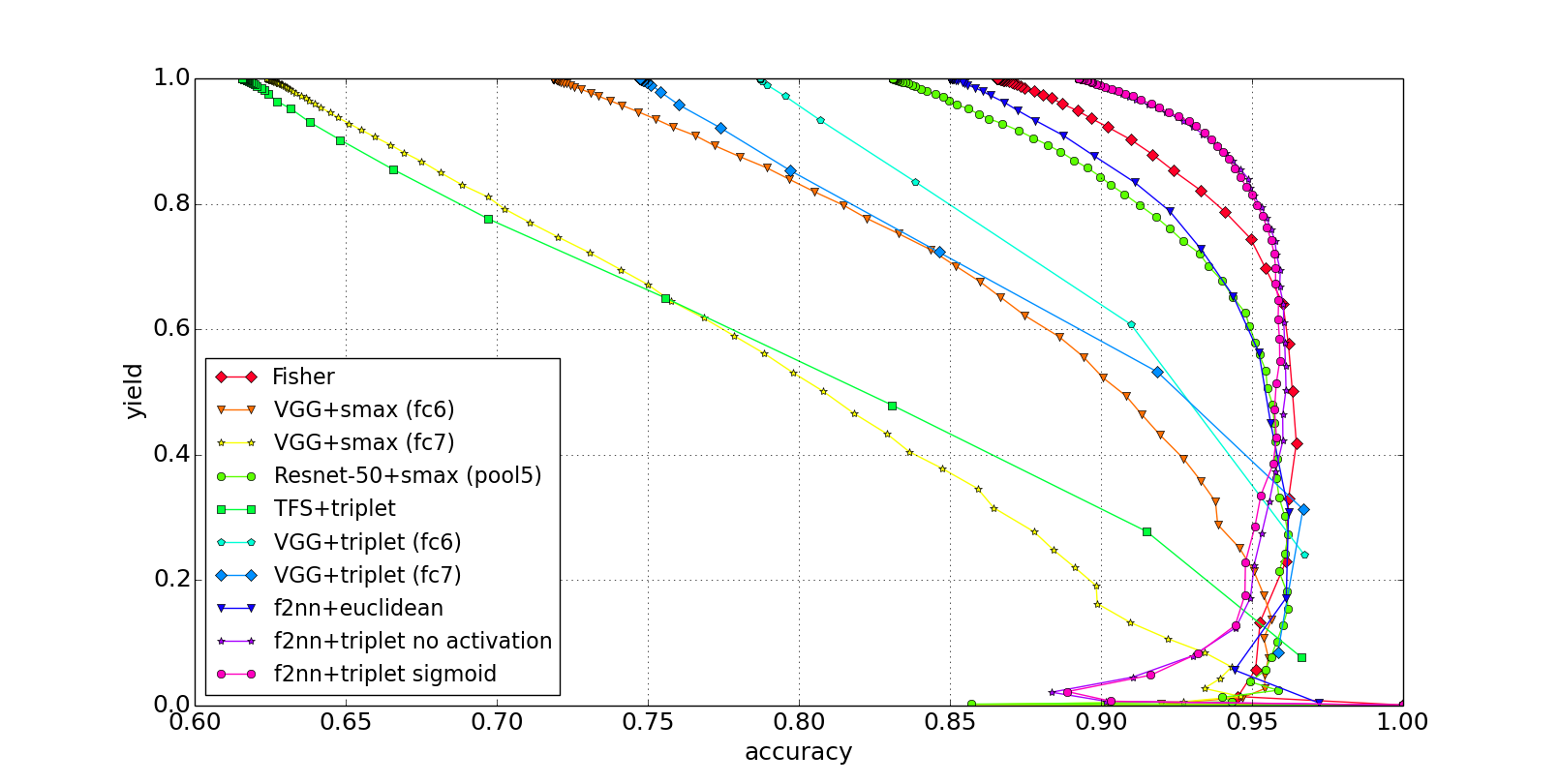}
            \caption{Yield vs Accuracy plot for the Malaysian Dataset}
            \label{fig:mhi2_acc}
        \end{center}
        \end{figure}
        
        \begin{figure}[!t]
        \begin{center}
            \includegraphics[width=1.1\linewidth]{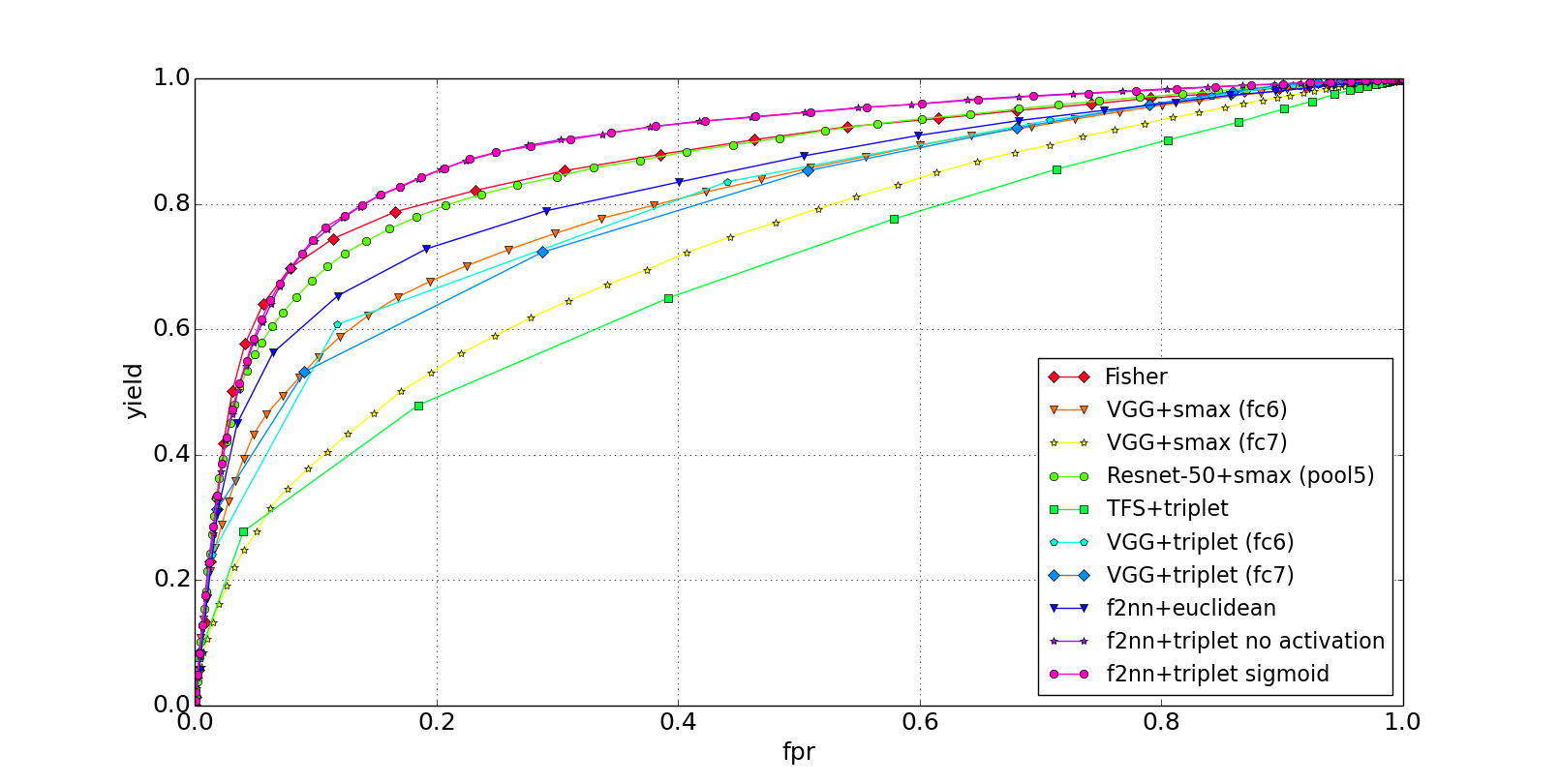}
            \caption{Yield vs FPR plot for the Malaysian Dataset}
            \label{fig:mhi2_fpr}
        \end{center}
        \end{figure}

        \begin{figure}[!t]
            \begin{center}
                \includegraphics[width=1.2\linewidth]{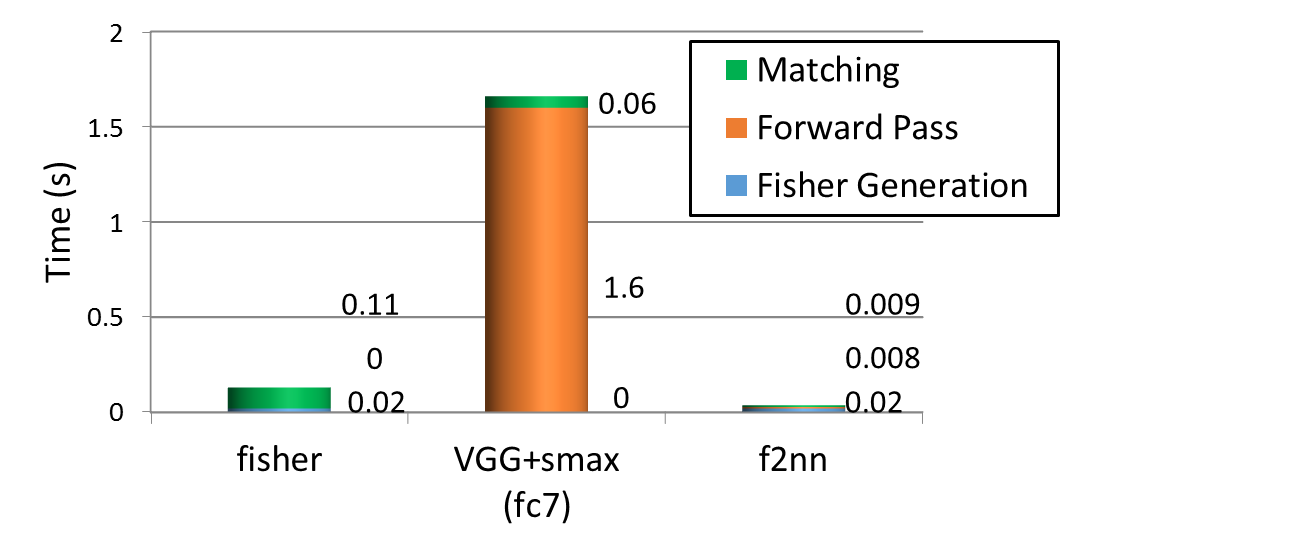}
                \caption{Processing time comparison of Fisher Vectors, VGG and \emph{f2nn} signatures}
                \label{fig:mta2_match}
            \end{center}
        \end{figure}   
\section{Results, Discussion, and Future Work}\label{sec:results}
    
    ROC curves for the two benchmark sets are shown in figures \ref{fig:mta2_acc} to \ref{fig:mhi2_fpr}.

    We find that while the CNN models perform very well (substantially better than the raw Fisher vector) on plates similar to the training set (even though the actual plates have not been seen before), their performance degrades on the Malaysian test set, to fall far behind even the raw Fisher vector. In contrast, the fine-tuned Fisher vector stays competitive with the CNNs on the American licence plates while not degrading on the Malaysian test set. Also, we find that the TFS model doesn't perform any better than the large, pre-trained models, and suffers quite heavily when faced with very different images such as the Malaysian benchmark set.
 
    In addition, we show the results of average time to process an image for unsupervised, supervised and hybrid methods in fig. (\ref{fig:mta2_match}). The time comparison is CPU only matching time for the American license plate transfer dataset containing 6006 templates. In this figure, "Generation" correspond to generation of Fisher Vectors while the “forward pass” corresponds to doing the forward pass over the net. The machine used for benchmarking is a 3GHz core machine with 32 GB RAM. 

    \subsection{Effect of the loss function \label{sec:results:loss_fn}}
        We find a wide variation in performance of the Fisher vector embedding across different loss functions used while training. While the autoencoder-based loss doesn't vary as much in performance, its improvement over the raw Fisher vector itself is quite minimal. The best results are seen with the triplet loss, which radically improves the performance of the embedding. That the triplet loss is the strongest factor in improving the embedding is seen even more clearly when we use a simple linear embedding (a single linear layer with no non-linearities afterward) of the Fisher vector, optimised under triplet loss. The results of such a model are close to the performance of the best Fisher embedding overall. The performance of \emph{f2nn} trained with Quadruplet loss (equation \ref{eq:quad}) was not at par with the \emph{f2nn} trained with the Triplet loss (equation \ref{eq:Trip}) and hence we do not report its performance. On the other hand, some loss functions such as the softmax, and consequently, the center loss couldn't be trained at all.
        This may be because our training dataset has much fewer images (around 100,000) and more individual classes (of the order of 40,000) than ImageNet.
        
        CNN performance also varies with the loss function. While we find very good performance of almost all the CNN models we tested on the American benchmark set (trained on any loss function), we see that their accuracy and yield drop off when applied to the Malaysian dataset, and the decline is more pronounced in the networks trained with triplet loss. This suggests that the triplet loss helps learn a good representation in domains similar to the training set, but could be suboptimal in adapting directly to a slightly more different domain. In the case of the Fisher vector, it is possible that the black-box vectors we inherited contained enough general ``knowledge'' about licence plates, such that the triplet loss would not ``specialize'' the representation to the American training set too much. In that case, we would expect the CNN performance to improve in the other domain if the pre-trained CNNs were exposed to more general licence plate images. This could be explored in more detail in the future. Also, similar to the Fisher vector, VGG net could not be trained with the center loss.
        
        The results suggest that it could be worth exploring Fisher vectors (and/or other unsupervised image representations) for their ability to represent general attributes of the training data such that they generalize well to unseen (but loosely similar) domains. Also, \emph{f2nn} signatures can be used over Deep Nets when time requirements are stringent. 
        
    We also did an error analysis at 100\% yield to get an insight into what was causing the \emph{f2nn} to fail on 2\% of American dataset. It was found that majority of the errors came because of the a bad cropped image being selected as template or test image. 
        Based on the error analysis, the next step would be to devise a methodology such that a good image is used as a template image in the database.
        
        Another interesting option would be to experiment with improved mining of triplets presented in \cite{kumar2017smart} where anchor and negative examples are assumed to be normally distributed and sampled accordingly.

\section*{Acknowledgements \label{sec:ack}}
    The authors would like to thank Manasa Kolla, Mayank Gupta, Rahul Mishra, Lalitha KS and Pragathi Praveena for their valuable inputs and fruitful discussions.


{\small
\bibliographystyle{ieee}
\bibliography{egbib}
}

\end{document}